%% file: paper.tex
\documentstyle[alltt,eepic]{article}

\gdef\SetFigFont#1#2#3#4#5{\it} % for compatibility with older versions of latex
\newcommand{\dashlinestretch}{30}

\def\la{\(\lambda\)}
\def\myand{\(\land\)}
\def\to{\(\rightarrow\)}
\def\ra{\(\rightarrow\)}
\def\so#1{\{#1\}}

\def\minus{\setminus}
\def\union{\cup}

\author{Marc Dymetman \quad Max Copperman \\
Rank Xerox Research Centre \\ 6, chemin de Maupertuis, Meylan 38240, France \\{\tt \{dymetman,copperman\}@xerox.fr}}

\title{Extended Dependency Structures and their Formal
  Interpretation \thanks{\  Copyright \copyright\  Xerox 1996}}

\date{}

\begin{document}

\maketitle

\bibliographystyle{plain}

\nocite{barwise-cooper:81,Melcuk:DepSynt}

\sloppy

\begin{abstract}
  We describe two ``semantically-oriented'' dependency-structure
  formalisms, U-forms and S-forms. U-forms have been previously used
  in machine translation as interlingual representations, but without
  being provided with a formal interpretation.  S-forms, which we
  introduce in this paper, are a scoped version of U-forms, and we
  define a compositional semantics mechanism for them. Two types of
  semantic composition are basic: complement incorporation and
  modifier incorporation. Binding of variables is done at the time of
  incorporation, permitting much flexibility in composition order and
  a simple account of the semantic effects of permuting several
  incorporations.
\end{abstract}

\fussy

\section{INTRODUCTION}

{\em U-forms} (Unscoped dependency form) are a representation
formalism which has been used (under a different name) as the basis
for the intermediary language in the machine translation system
CRITTER \cite{IsDyMa88,These-Dym,IsabelleDoc}. U-forms account for two
central aspects of linguistic structure: predicate-argument relations
and headedness (complements vs.  modifiers), and so form a middle
ground between a ``semantic'' and a ``syntactic'' representation.
This, combined with their formal simplicity, accounts for much of the
popularity of U-forms or related formalisms --- such as the semantic
and deep syntactic representations used in Mel'cuk's Meaning-Text
Theory \cite{melcuk81} --- in applications such as machine translation
and text generation.

Although U-forms are strongly ``meaning-oriented'', their
interpretation is never made explicit but is left to the computational
linguist's intuition. This has two consequences:
\begin{itemize}
\item Operations performed on U-forms and related formalisms cannot be
  controlled for semantic validity.  So, for instance, it is common
  practice to define graph rewriting rules on these representations
  which are believed to produce semantically equivalent expressions.
  Without the check of formal interpretation, these rules may work in
  some cases, but produce wrong results in other cases. So for
  instance, a rule rewriting (the representation of) ``John's salary
  is \${}25000 higher this year than last year'' into ``John's salary
  was \${}25000 lower last year than this year'' would seem
  intuitively valid until one considered the case of ``John's salary
  is 50\% higher this year than last year'', where it does not work
  any more.
\item U-forms are not directly adapted to applications putting
  emphasis on denotational semantics and formal reasoning, like for
  instance some
  natural language generation systems in well-formalized domains
  \cite{Huang,Ranta,Levine}, see also \cite{Alshawi:CLE}.
\end{itemize}

A basic obstacle to providing a formal interpretation for U-forms is
the fact that these representations leave the relative scopes of
dependents implicit. The S-form representation (Scoped dependency
form), which we introduce here, is an extension of U-form notation
which makes scope explicit, by allowing dependents to be ordered
relative to one another. Dependents (complements or modifiers) can
move freely relative to one another in the S-form structure, under
certain binding-site constraints.

We then go on to provide a compositional interpretation mechanism for
S-forms. Free variables (generalizations of the arg$_1$, arg$_2$,
arg$_3$ annotations of standard dependency formalisms) are used to
connect an argument to its binding-site inside a predicate. Binding of
variables is done at the time of incorporation, permitting much
flexibility in composition order and a simple account of the semantic
effects of permuting several incorporations. This liberal use of free
variables is contrasted to the approach of Montague grammar, where the
requirement that semantic expressions entering into a composition are
closed (do not contain free variables) leads to a certain rigidity in
the order of composition.

Two kinds of semantic composition are basic: complement incorporation,
where the complement fills a semantic role inside the head, and
modifier incorporation, where the head fills a semantic role inside
the modifier. The mechanism of actually deriving the semantic
translation of the composition from the semantic translations of its
two components is handled through a list of type-sensitive composition
rules, which determine the action to be taken on the basis of the
component types. The flexibility of the approach is illustrated on an
example involving proper names, quantified noun phrases, adverbials
and relative clauses.

\section{U-FORMS}

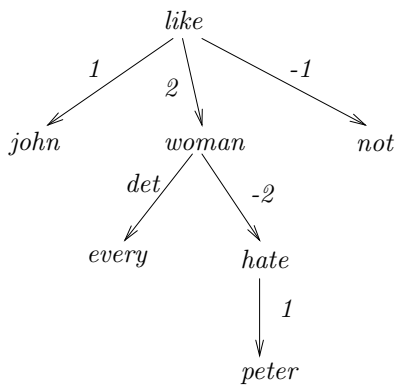
\begin{figure}[h]
  \begin{center}
    \leavevmode
    ~ \input{AT1.tex}
    \caption{A U-form.}
    \label{@1}
  \end{center}
\end{figure}

Formally, U-forms are unordered labelled n-ary trees such as the one
shown in Fig.~\ref{@1}, corresponding to the sentence: (S1) ``John does not
like every woman hated by Peter''.

\noindent

The edge labels are members of the set \{det, 1, 2, 3, ..., -1, -2,
-3, ...\}, and correspond either to determiners (label `det') or to
argument positions relative to a predicate node (other labels).

The U-form of Fig.~\ref{@1} expresses three predicate-argument relations
among the nodes:

\begin{figure}[h]
  \begin{center}
    \leavevmode
    ~ \input{D2.tex}
    \caption{Predicate-argument relations in a U-form.}
    \label{D2}
  \end{center}
\end{figure}
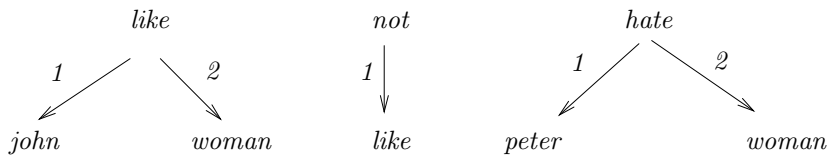

In order to extract the predicate-argument relations encoded into
the U-form, one needs to apply the following ``rule''. Let's notate
(A,L,B) an edge of the tree, where A is the upper vertex, B the lower
vertex, and L the edge label. With each node A in the tree, one
associates its set of {\em predication edges}, that is the set PA$_A$ of
edges of the form (A,+i,X) or (X,-i,A). One then considers the {\em
predication tree} T$_A$ made by forming the collection of edges (A,L,X)
where L is positive and either (A,L,X) or (X,inverse(L),A) is a
predication edge of A. Each predication tree denotes a
predicate-argument relation among U-form nodes. So for instance, the
tree T$_{\mbox{\scriptsize hate}}$ is formed by forming the edges (hate,1,peter) and
(hate,2,woman), and this corresponds to the predicate-argument
relation hate(peter,woman).

\paragraph{WELL-FORMEDNESS CONDITIONS ON U-FORMS}

In order to be well-formed, a U-form UF has to respect the following
condition. For any node A of UF, the predication tree T$_A$ must be such
that:
\begin{enumerate}
        \item {[{\em No holes condition}]} If (A,i,B) is an edge of T$_A$, then for
        any number j between 1 and i, T$_A$ must contain a node of form
        (A,j,C).

        \item {[{\em No repetition condition}]} No two edges of T$_A$ can have the
        same label i.
\end{enumerate}

\paragraph{MORE ON U-FORMS}

Negative labels are a device which permits to reconcile the notation
of predicate-argument structure with the notation of syntactic
dependency. So, in the U-form considered above, while ``semantically''
the `woman' node is an argument of the `hate' node, ``syntactically'' the
`hate' node is a dependent of the `woman' node. Cases such as this one,
where there is a conflict between predicate-argument directionality
and dependency directionality are notated in the U-form through
negative labels, and correspond to {\em modifiers}. Cases where the
directionality is parallel correspond to {\em complements}.

When used as interlingual representations in machine translation
systems, U-forms have several advantages. The first is that they
neutralize certain details of syntactic structure that do not
carry easily between languages. For instance, French and English
express negation in syntactically different ways: ``Rachel does not
like Claude'' vs. ``Rachel n'aime pas Claude''; this difference is
neutralized in the U-form representation, for both negations are
expressed through a single negation predicate in the U-form.

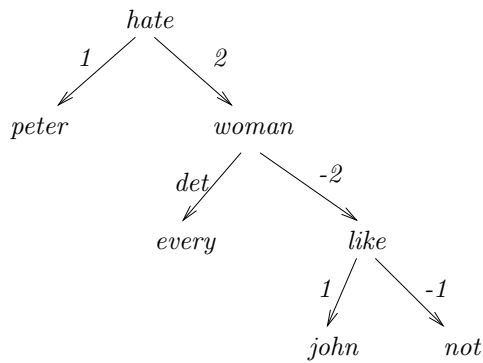
\begin{figure}[h]
  \begin{center}
    \leavevmode 
    ~ \input{D3.tex}
    \caption{A different U-form.}
    \label{D3}
  \end{center}
\end{figure}

A second advantage is that they represent a good compromise between
paraphrasing potential and semantic precision. So, for instance, in
the CRITTER system, the three sentences:

\begin{alltt}\sf
  John does not like every woman that Peter hates
  John does not like every woman hated by Peter
  Every woman whom Peter hates is not liked by John
\end{alltt}

\noindent would be assigned the U-form of Fig.~\ref{@1}. On the other
hand, the sentence:

\begin{alltt}\sf
  Peter hates every woman that John does not like
\end{alltt}

\noindent would be assigned the U-form of Fig.~\ref{D3}, which is
different from the previous U-form, although the predicate-argument
relations are exactly the same in both cases.

One can take advantage of such paraphrasing potential in certain cases
of syntactic divergence between languages. For instance, French does
not have a syntactic equivalent to the dative-movement + passive
configuration of:

\begin{alltt}\sf
  Rachel was given a book by Claude
\end{alltt}

\noindent so that a direct syntactic translation is not possible. However, at
the level of U-form, this sentence is equivalent to the French
sentence:

\begin{alltt}\sf
  Claude a donn\'e un livre \`a Rachel
\end{alltt}

\noindent and this equivalence can be exploited to provide a translation of the
first sentence.

One serious problem with U-forms, however, is that they do not have
unambiguous readings in cases where the relative scopes of
constituents can result in different semantic interpretations. So, in
the case of sentence (S1), the two readings: ``it is not the case that
John likes every woman hated by Peter'', and ``John dislikes every woman
that Peter hates'' are not distinguished by the U-form
of Fig.~\ref{@1}. 

\section{S-FORMS}

\paragraph{INTRODUCING SCOPE}

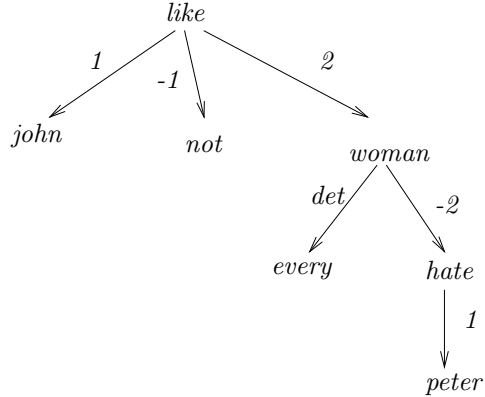
\begin{figure}[h]
  \begin{center}
    \leavevmode
    ~ \input{AT2.tex}
    \caption{Introducing scope by ordering the nodes.}
    \label{@2}
  \end{center}
\end{figure}

Let's consider the tree represented in Fig.~\ref{@2}. 

The only difference between this tree and the U-form of Fig.~\ref{@1} is
that the nodes of our new tree are considered ordered whereas they
were considered unordered in the U-form. The convention is now that
dependent sister nodes are interpreted as having different scopes,
with narrower scope corresponding to a position more to the right.

The tree of Fig.~\ref{@2} can be glossed in the following way:

\begin{alltt}\sf
  John, it is not the case that he likes every woman that Peter hates
\end{alltt}

If we consider the six permutations of the nodes under like, we can
produce six different scopings.
Because John refers to an individual, not a quantified NP, these six
permutations really correspond to only the two interpretations
given above.  The tree of
Fig.~\ref{@2} corresponds to the first of these interpretations, which is
the preferred interpretation for sentence (S1).

Our discussion of scope being represented by node order has been
informal so far. In order to make it formal, we need to encode our
representation into a binary-tree format on which a compositional
semantics can be defined. To do that, in a first step we replace the
argument numbers of Fig.~\ref{@2} by explicit argument names; in a second
step we encode the resulting ordered n-ary tree into a binary format
which makes explicit the order in which dependents are incorporated
into their head.

\paragraph{S-FORMS}

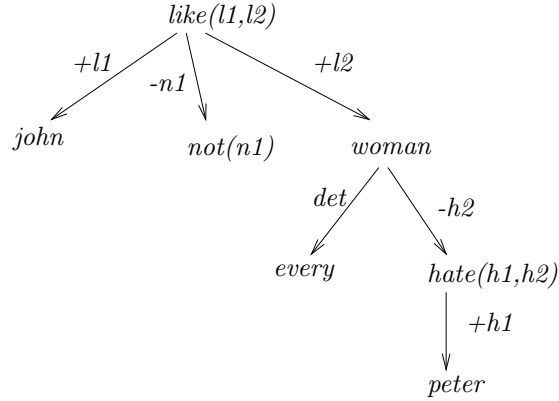
\begin{figure}[h]
  \begin{center}
    \leavevmode
    ~ \input{AT4.tex}
    \caption{An S-form.}
    \label{@4}
  \end{center}
\end{figure}

Consider the n-ary tree of Fig.~\ref{@2}. For any node A in this tree,
take the set of predication edges associated with A, that is the set
of edges (A,+i,B$_i$) and (B$_i$,-i,A). By renaming each such node A
into A(X$_1$,..,X$_n$), where X$_1$,...,X$_n$ are fresh identifiers,
and by renaming each such label +i (resp. -i) into +X$_i$ (resp.
-X$_i$), one obtains a new tree where argument numbers have been
replaced by argument names. For instance the previous representation
now becomes the tree of Fig.~\ref{@4}.

\noindent This representation is called a {\em scoped dependency form}, or {\em
S-form}.

\paragraph{BINARY TREE ENCODING OF S-FORMS: B-FORMS}

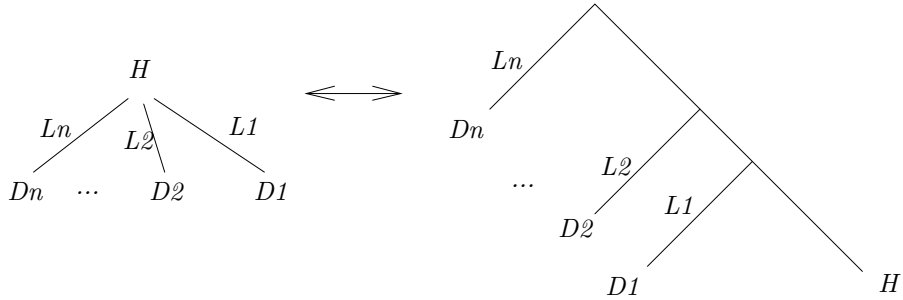
\begin{figure}[h]
  \begin{center}
    \leavevmode
    ~ \input{D5.tex}
    \caption{The transformation between S-forms and B-forms.}
    \label{D5}
  \end{center}
\end{figure}

In order to encode the ordered n-ary tree into a binary tree, we
need to apply recursively the transformation illustrated in Fig.~\ref{D5},
which consists in forming a ``head-line'', projecting in a north-west
direction from the head H, and in ``attaching'' to this line
''dependent-lines'' D$_1$, D$_2$, ..., D$_n$, with D$_1$ the rightmost dependent
(narrowest scope) and D$_n$ the leftmost dependent (widest scope) in the
original tree.

\begin{figure}[h]
  \begin{center}
    \leavevmode
    ~ \input{AT6.tex}
    \caption{A B-form.}
    \label{@6}
  \end{center}
\end{figure}
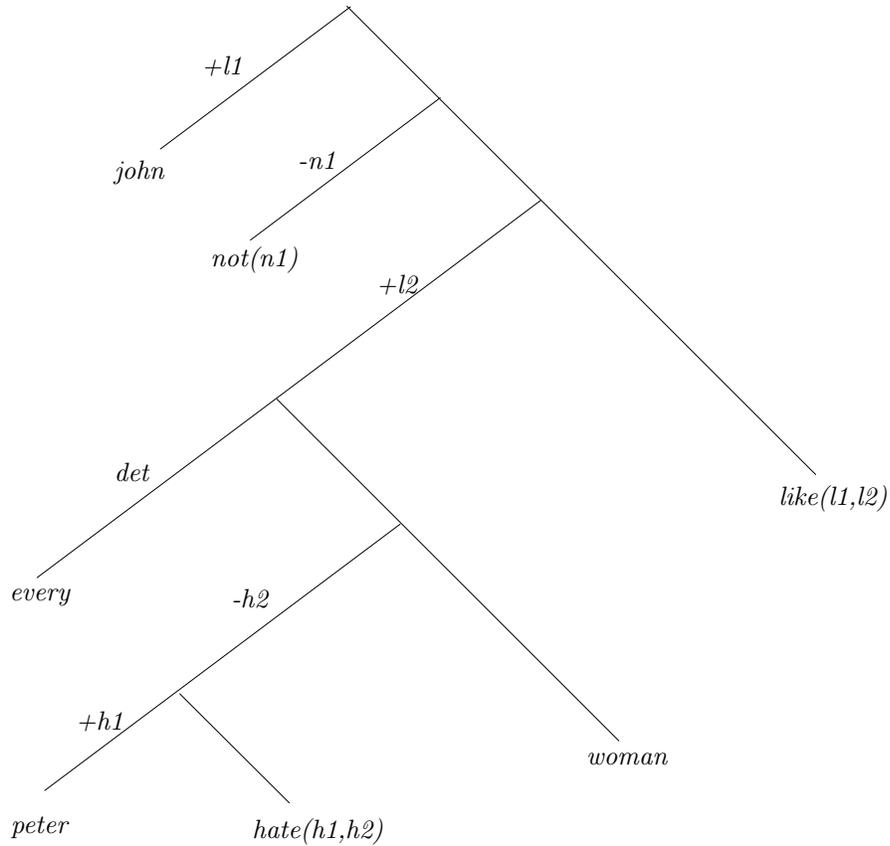

Applying this encoding to our example, we obtain the binary tree of
Fig.~\ref{@6}, which is called a {\em B-form}.

The B-form makes explicit the order of incorporation of dependents into
the head-line. By permuting several dependent-lines along their
head-line, this incorporation order is changed and gives rise to
different scopings.

{\bf S-forms and B-forms are completely equivalent representations}.
Clearly, the encoding, called the {\em S-form/B-form encoding}, which
has just been defined is reversible. The S-form is more compact and
makes the dependency relations more conspicuous, whereas the B-form
makes the compositionality more explicit.

\paragraph{WELL-FORMEDNESS CONDITIONS ON B-FORMS AND S-FORMS}

Starting from the U-form and enriching it, we have informally
introduced the notions of S-form and B-form. We now define them
formally.

We start by giving a recursive definition of IBFs (incomplete
B-forms), that is, B-forms which may contain unresolved free
variables. We use the notation ((D,Label),H) the labelled binary tree
obtained by taking H as the right subtree, D as the left subtree, and
by labelling the left edge with Label. We also use the notation
fv(IBF) for the set of the free variables in IBF.

\subparagraph{DEFINITION OF INCOMPLETE B-FORMS}

\sloppy
\begin{enumerate}
\item A node N of the form Pred(x1,..,xn)
  is an IBF with the set of free variables fv(N) =
  \{x1,..,xn\};

  \item If D and H are IBFs, fv(D) and fv(H) are disjoint, and 
  x \(\in\) fv(H) then H'=((D,+x),H) is an IBF with 
  fv(H') = fv(D) \(\union\) fv(H) \(\minus\) \{x\};

  \item If D and H are IBFs, fv(D) and fv(H) are disjoint, and x \(\in\) fv(D) 
  then H'=((D,-x),H) is an IBF with fv(H') = fv(D) \(\union\) fv(H) \(\minus\) \{x\};

  \item If D and H are IBFs, and fv(D) and fv(H) are disjoint, then
  H'=((D,det),H) is an IBF with fv(H') = fv(D) \(\union\) fv(H). 
\end{enumerate}
\fussy

\subparagraph{DEFINITION OF B-FORMS} A B-form is an IBF with an empty
set of free variables.

The notion of S-form can now be defined through the use of the
S-form/B-form encoding.

\subparagraph{DEFINITION OF S-FORMS} A S-form is an ordered labelled
n-ary tree which can be obtained from a B-form through the inverse
application of the S-form/B-form encoding.

\medskip

It can be easily verified that the representation of Fig.~\ref{@6} is
indeed a B-form, and, consequently, the representation of
Fig.~\ref{@4} is a valid S-form. More generally, it can be easily
verified that enriching a U-form by ordering its nodes, and then
replacing argument variables by argument names always results in a
valid S-form.\footnote{The converse is not true: not all S-forms can
  be obtained in this way from a U-form. For instance, there exists a
  S-form corresponding to the preferred reading for ``Fido visited
  most trashcans on every street'', which has ``every street''
  outscoping ``most trashcans'', and which is not obtained from a
  U-form in this simple way. However, there exists a mapping from
  S-forms to U-forms, the {\em scope-forgetting mapping}, which
  permits to define equivalence classes among S-forms ``sharing'' the
  same U-form. This relation between S-forms and U-forms can be used
  to give a (non-deterministic) formal interpretation to U-forms, by
  considering the interpretations of the various S-forms associated
  with it (see the technical report companion to this paper.)}

\section{THE INTERPRETATION PROCESS}

We now describe the interpretation process on B-forms. Interpretation
proceeds by propagating semantic translations and their types
bottom-up.

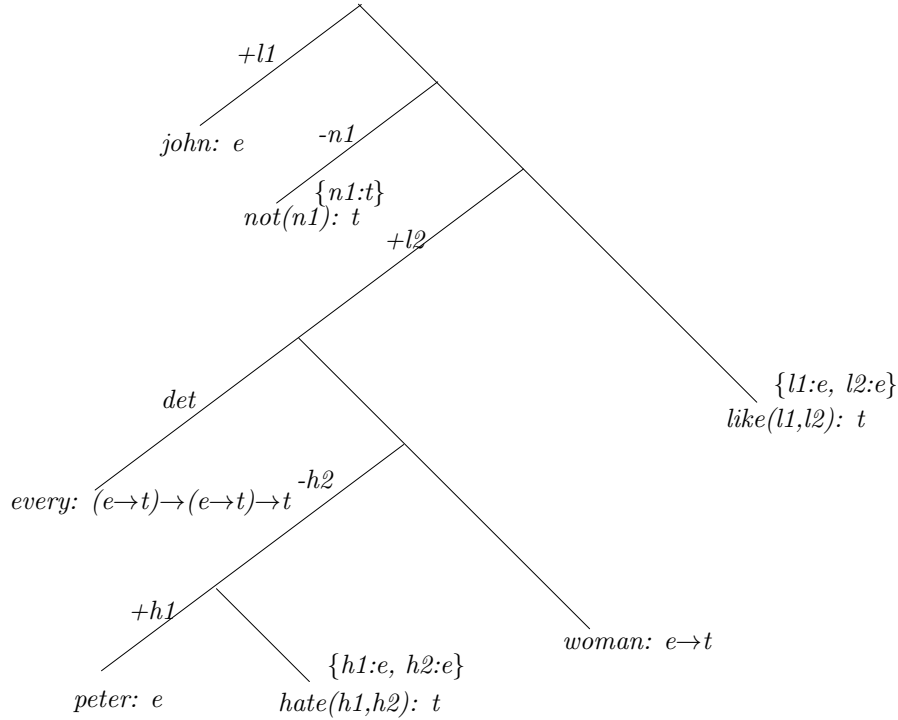
\begin{figure}[h]
  \begin{center}
    \leavevmode
    ~ \input{AT7.tex}
    \caption{Typing the leaves. The free variables and their types are
      indicated in brackets.}
    \label{@7}
  \end{center}
\end{figure}

The first step consists in typing the leaves of the tree, while
keeping track of the types of free variables, as in Fig.~\ref{@7}.

The types given to the leaves of the tree are the usual functional
types formed starting with e (entities) and t (truth values). In the
case where the leaf entity contains free variable arguments, the types
of these free variables are indicated, and the type of the leaf takes
into account the fact that these free variables have already been
included in the functional form of the leaf. Thus hate(h1,h2), which
can be glossed as: ``h1 hates h2'', is given type t, while h1 and h2 are
constrained to be free variables of type e.

\paragraph{VARIABLE-BINDING RULES}

\sloppy

According to the well-formedness conditions for B-forms, a complement
incorporation ((D,+x),H) is only possible when H contains x among its
free variables; the ``syntactic dependent'' D is seen as semantically
``filling'' the place that x occupies in the ``syntactic head'' H. In the
same way, a modifier incorporation ((D,-x),H) is only possible when D
contains x among its free variables; in this case the ``syntactic'' head
H is seen as semantically ``filling'' the place that x occupies in the
``syntactic dependent'' D. (This difference corresponds to the
opposition which is sometimes made between syntactic and semantic
heads and dependents: complements are dependents both syntactically
and semantically, while modifiers are syntactically dependents but
semantically heads.) 

\fussy

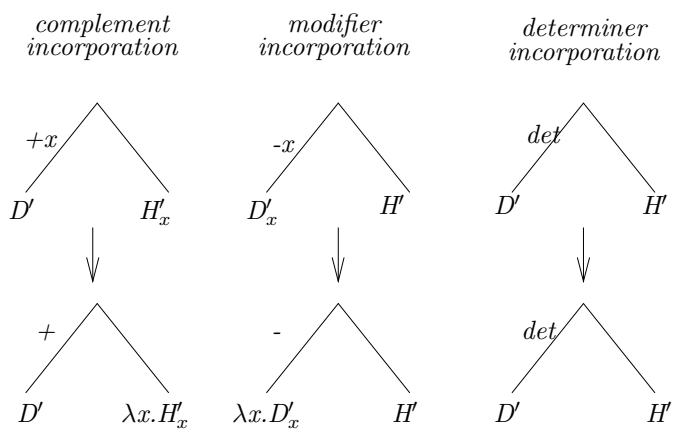
\begin{figure}[h]
  \begin{center}
    \leavevmode
    ~ \input{AT8.tex}
    \caption{Variable-binding rules. D' and H' correspond to the
      semantic translation of the subtrees rooted in D and H respectively.}
    \label{@8}
  \end{center}
\end{figure}

In order to make formal sense of the informal notion ``filling the
place of x in A$_x$'' (where the notation A$_x$ means that A contains the
free variable x), we introduce the variable-binding rules of Fig.~\ref{@8}. 

\def\pp{{+}}
\def\mm{{-}}
\def\px{{+x}}
\def\mx{{-x}}
\def\dprim{D\('\)}
\def\hprim{H\('\)}
\def\dprimx{D\('_x\)}
\def\hprimx{H\('_x\)}
\def\ldprim{\la{}x.D\('_x\)}
\def\lhprim{\la{}x.H\('_x\)}

These rules tell us how to ``get rid'' of the free variable being bound
during complement or modifier incorporation, namely by forming the
abstraction \la{}x.A$_x$ before actually performing the semantic
composition between the dependent and the head. For completeness,
determiner incorporation, which does not involve variable binding, is
given along with complement and modifier incorporation.

Two things should be noted about this way of ``delaying''
variable-binding until the relevant dependent is incorporated:
\begin{itemize}
\item Suppose that we had bound the variables appearing in the head
  predicate locally, that is to say, that, in the style of Montague
  grammar \cite{Gamut:Logic2}, we had written \la{}l2l1.like(l1,l2) 
  instead of like(l1,l2), and so forth, in
  Fig.~\ref{@6}. Then each incorporation of a dependent into the
  ``head-line'' would have changed the type of the head; thus `not'
  would have had to combine either with a head of type e\to{}e\to{}t,
  or e\to{}t, or t, depending on its scope relative to the other
  dependents; with the scheme adopted here, the type of the head
  remains invariant along the head-line;
\item Under the same hypothesis, the incorporation of the second
  argument first and of the first argument second would have been much
  simpler than the reverse incorporation order, and some mechanism
  would have had to be found to distinguish the two orders. Then
  permuting the relative order of two dependents along the head-line
  --- corresponding to different scope possibilities --- would have
  had complex computational consequences. In the scheme adopted here,
  these cases are handled in a uniform way.
\end{itemize}

The way free variables are used in our scheme is somewhat reminiscent
of the use of {\em syntactic variables} {he$_n$} in Montague
grammar. Montague grammar has the general requirement that only closed
lambda-terms (lambda terms containing only bound variables) are
composed together. This requirement, however, is difficult to
reconcile with the flexibility needed for handling quantifier scope
ambiguities. Syntactic variables are a device which permit to
``quantify into'' clauses at an arbitrary time, bypassing the normal
functional composition of lambda-terms, which requires a strict
management of incorporation order. In our scheme, by contrast, this
secondary mechanism of Montague grammar is graduated to a central
position. Composition is always done between two lambda-terms one of
which at least contains a free variable which gets bound at the
time of incorporation.

\paragraph{TYPE SENSITIVE COMPOSITION RULES}

\sloppy

If we apply the variable-binding rules to the subtree PH =
((peter,-h1),hate(h1,h2)) of Fig.~\ref{@7}, we find that we must compose the
semantic translations peter and \la{}h1.hate(h1,h2) in
``complement'' (+) mode. The first function is of type e, while the
second function is of type e\to{}t (for hate(h1,h2) is of type t, and
h1 of type e).

\fussy

How do we compose two such functions? A first solution, in the spirit
of Lambek calculus \cite{Morrill:Type-Logical} or of linear logic
\cite{DLPS:QuantLFG}, would be to define a general computational
mechanism which would be able, through a systematic discipline of
type-changing operations, to ``adapt'' automatically to the types of
the functions undergoing composition.

Such mechanisms are powerful, but they tend to be algorithmically
complex, to be non-local, and also to give rise to spurious
ambiguities (superficial variations in the proof process which do not
correspond to different semantic readings). 

Here, we will prefer to use a less general mechanism, but one which
has two advantages. First, it is local, simple, and efficient. Second,
it is flexible and can be extended to handle the semantics of
sentences extracted from a real corpus of texts, which it might be
perilous to constrain too strongly from the start.

The mechanism is the following. We establish a list of acceptable
``type-sensitive composition rules'', which tell us how to compose two
functions according to their types. Such a (provisory) list is given
below:\footnote{It is a matter for further research to propose
  principles for producing such rules. Some of them can be seen as
  special cases of general type-raising principles, others (such as
  C5) are necessary if one accepts that the type of intersective
  adjectives and restrictive relative clauses has to be e\to t.}

\begin{alltt}
(C1) composition(+, L:T->S, R:T, L(R):S)
(C2) composition(+, L:e, R:e->t, R(L):t)
(C3) composition(det, L:T->S, R:T, L(R):S)
(C4) composition(-, L:T->S, R:T, L(R):S)
(C5) composition(-, L:e->t, R:e->t,\la{}x.R(x)\myand{}L(x):e->t)
    ...
\end{alltt}

The entries in this list have the following format. The first argument
indicates the type of composition (`+' for complement incorporation,
`-' for modifier incorporation, `det' for determiner incorporation);
the second argument is of the form Left:LeftType, where Left is the
left translation entering the composition, and LeftType is its type;
similarly, the second argument Right:RightType corresponds to the
right subtree entering the composition; finally the third argument
gives the result Result:ResultType of the composition, where the
notation A(B) has been used to indicate standard functional
application of function A on argument B. Uppercase letters indicate
unifiable variables.

It may be remarked that if, in these rules, we neglect the functions
themselves (Left, Right, Result) and concentrate on their types
(LeftType, RightType, ResultType), then the rules can be seen as
imposing constraints on what can count as validly typed trees; these
constraints can flow from mother to daugthers as well as in the
opposite direction. Thus, through these rules, knowing that the
head-line functions projecting from a verbal head must be of type t
imposes some constraints on what are the possible types for the
dependents; this can be useful in particular for constraining the
types of semantically ambiguous lexical elements.

\def\hateone{\it hate(peter,h2): t}
\def\womanone{\it \la h2.woman(h2)\myand hate(peter,h2): e\to t}
\def\womantwo{\it \la P.every(\la h2.woman(h2)\myand hate(peter,h2),P):
  (e\to t)\to t}
\def\likeone{\it every(\la h2.woman(h2)\myand hate(peter,h2),\la
  l2.like(l1,l2)): t}
\def\liketwo{\it not(every(\la h2.woman(h2)\myand hate(peter,h2),\la
  l2.like(l1,l2))): t}
\def\likethree{\it not(every(\la h2.woman(h2)\myand hate(peter,h2),\la
  l2.like(john,l2))): t}

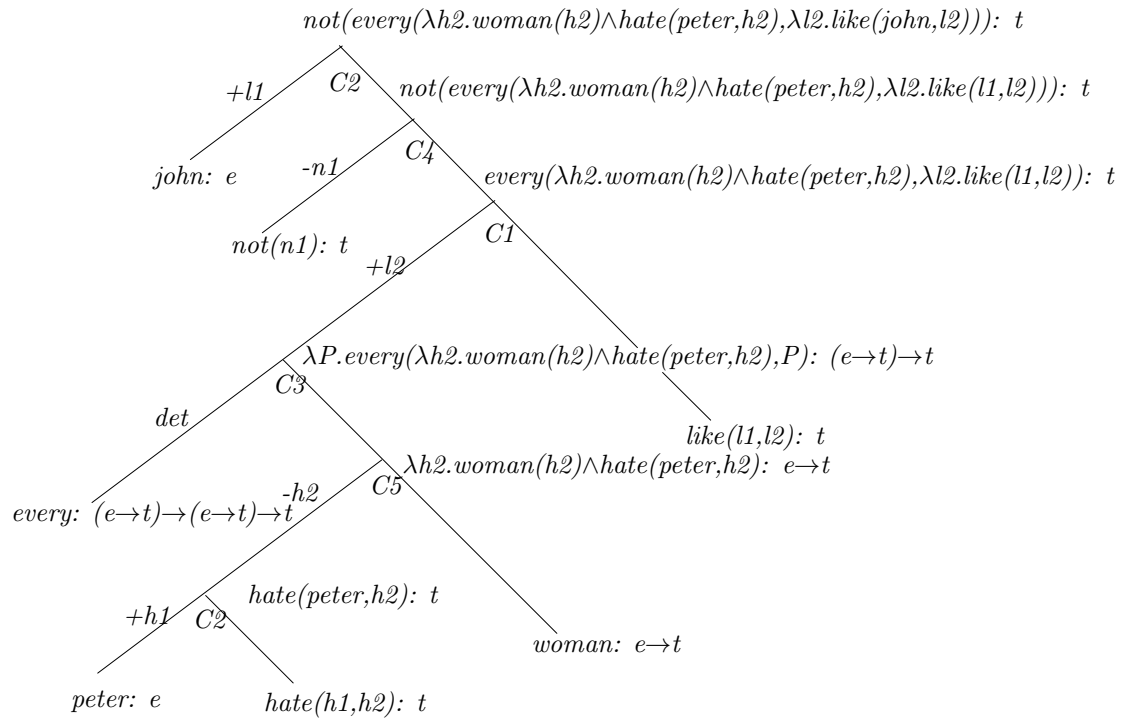
\begin{figure}[p]
  \begin{center}
    \leavevmode
    \hspace*{-1cm} \input{AT9.tex}
    \caption{B-form interpretation. For `every', we make use of the generalized quantifier
      notation {\em quant(restriction,scope)}.}
    \label{@9}
  \end{center}
\end{figure}

If we now go back to our example, we have to compose in
complement mode (+) the function peter, of type e, with the function
\la{}h1.hate(h1,h2), of type e\to{}t. Consulting the list of
composition rules, we see that the only applicable rule is (C2), and
that the result is \la{}h1.hate(h1,h2) (peter) = hate(peter,h2),
of type t.

\begin{sloppypar}
Now that we have the semantic translation hate(peter,h2) for the
subtree PH, we can compute the translation for the subtree
((PH,-h2),woman). By the variable-binding rule for modifiers, we need
first to form the abstraction \la{}h2.hate(peter,h2), of type
e\to{}t, and compose it in `-' mode with woman, of type
e\to{}t. Consulting the list of composition rules, we find that the
only applicable rule is (C5), and that the result of this application
is \la{}h2.woman(h2)\myand{}hate(peter,h2).\footnote{The rule (C5)
  differs from the previous rules in the list in that it introduces
  the logical connective \myand{} which does not originate in
  functional material already present in either of the arguments. A
  possible justification for the rule, however, is that it allows
  conferring the ``natural'' type e\to{}t to an (intersective) adjective
  such as ``black'', or for a relative modifier such as ``hated by
  peter'', and also that there does not seem to exist any good reason
  why type composition should be restricted to ``functionally matching''
  types only. Semantic type coercions abound in natural language, as
  in the case of ``glass elephant'', ``short win'', etc., and these
  require complex composition operations on the elements combined.}
\end{sloppypar}

The process of semantic translation proceeds in this way bottom-up on
the B-form. The end result is shown in Fig.~\ref{@9}.

\subsubsection*{Acknowledgments} Thanks to Alain Lecomte and Fr\'ed\'erique
Segond for comments and discussions.

% \bibliography{coling96}

\end{document}

%% file: AT1.tex
\setlength{\unitlength}{0.00066667in}
\begingroup\makeatletter\ifx\SetFigFont\undefined%
\gdef\SetFigFont#1#2#3#4#5{%
  \reset@font\fontsize{#1}{#2pt}%
  \fontfamily{#3}\fontseries{#4}\fontshape{#5}%
  \selectfont}%
\fi\endgroup%
{\renewcommand{\dashlinestretch}{30}
\begin{picture}(2930,2883)(0,-10)
\path(1275,2685)(300,2010)
\path(381.587,2102.971)(300.000,2010.000)(415.739,2053.639)
\path(1350,2685)(1500,2010)
\path(1444.683,2120.635)(1500.000,2010.000)(1503.254,2133.650)
\path(1500,2685)(2775,2010)
\path(2654.909,2039.633)(2775.000,2010.000)(2682.982,2092.660)
\path(1425,1785)(900,1110)
\path(949.992,1223.140)(900.000,1110.000)(997.353,1186.304)
\path(1500,1785)(1950,1110)
\path(1858.474,1193.205)(1950.000,1110.000)(1908.397,1226.487)
\path(1950,810)(1950,210)
\path(1920.000,330.000)(1950.000,210.000)(1980.000,330.000)
\put(1200,2760){\makebox(0,0)[lb]{\smash{{{\SetFigFont{10}{12.0}{\rmdefault}{\mddefault}{\itdefault}like}}}}}
\put(1800,30){\makebox(0,0)[lb]{\smash{{{\SetFigFont{10}{12.0}{\rmdefault}{\mddefault}{\itdefault}peter}}}}}
\put(600,2385){\makebox(0,0)[lb]{\smash{{{\SetFigFont{10}{12.0}{\rmdefault}{\mddefault}{\itdefault}1}}}}}
\put(2175,2385){\makebox(0,0)[lb]{\smash{{{\SetFigFont{10}{12.0}{\rmdefault}{\mddefault}{\itdefault}-1}}}}}
\put(900,1485){\makebox(0,0)[lb]{\smash{{{\SetFigFont{10}{12.0}{\rmdefault}{\mddefault}{\itdefault}det}}}}}
\put(1875,1410){\makebox(0,0)[lb]{\smash{{{\SetFigFont{10}{12.0}{\rmdefault}{\mddefault}{\itdefault}-2}}}}}
\put(2100,510){\makebox(0,0)[lb]{\smash{{{\SetFigFont{10}{12.0}{\rmdefault}{\mddefault}{\itdefault}1}}}}}
\put(0,1815){\makebox(0,0)[lb]{\smash{{{\SetFigFont{10}{12.0}{\rmdefault}{\mddefault}{\itdefault}john}}}}}
\put(1200,1815){\makebox(0,0)[lb]{\smash{{{\SetFigFont{10}{12.0}{\rmdefault}{\mddefault}{\itdefault}woman}}}}}
\put(2700,1808){\makebox(0,0)[lb]{\smash{{{\SetFigFont{10}{12.0}{\rmdefault}{\mddefault}{\itdefault}not}}}}}
\put(600,938){\makebox(0,0)[lb]{\smash{{{\SetFigFont{10}{12.0}{\rmdefault}{\mddefault}{\itdefault}every}}}}}
\put(1800,893){\makebox(0,0)[lb]{\smash{{{\SetFigFont{10}{12.0}{\rmdefault}{\mddefault}{\itdefault}hate}}}}}
\put(1200,2235){\makebox(0,0)[lb]{\smash{{{\SetFigFont{10}{12.0}{\rmdefault}{\mddefault}{\itdefault}2}}}}}
\end{picture}
}

%% file: D2.tex
\setlength{\unitlength}{0.00069991in}
\begingroup\makeatletter\ifx\SetFigFont\undefined%
\gdef\SetFigFont#1#2#3#4#5{%
  \reset@font\fontsize{#1}{#2pt}%
  \fontfamily{#3}\fontseries{#4}\fontshape{#5}%
  \selectfont}%
\fi\endgroup%
{\renewcommand{\dashlinestretch}{30}
\begin{picture}(6028,1059)(0,-10)
\path(4695,824)(4102,291)
\path(4171.193,393.529)(4102.000,291.000)(4211.302,348.906)
\path(4792,846)(5550,321)
\path(5434.270,364.663)(5550.000,321.000)(5468.432,413.988)
\put(4590,936){\makebox(0,0)[lb]{\smash{{{\SetFigFont{10}{12.0}{\rmdefault}{\mddefault}{\itdefault}hate}}}}}
\put(5490,36){\makebox(0,0)[lb]{\smash{{{\SetFigFont{10}{12.0}{\rmdefault}{\mddefault}{\itdefault}woman}}}}}
\put(3690,36){\makebox(0,0)[lb]{\smash{{{\SetFigFont{10}{12.0}{\rmdefault}{\mddefault}{\itdefault}peter}}}}}
\put(4185,614){\makebox(0,0)[lb]{\smash{{{\SetFigFont{10}{12.0}{\rmdefault}{\mddefault}{\itdefault}1}}}}}
\put(5250,636){\makebox(0,0)[lb]{\smash{{{\SetFigFont{10}{12.0}{\rmdefault}{\mddefault}{\itdefault}2}}}}}
\path(900,711)(225,261)
\path(308.205,352.526)(225.000,261.000)(341.487,302.603)
\path(1125,711)(1575,261)
\path(1468.934,324.640)(1575.000,261.000)(1511.360,367.066)
\path(2797,809)(2797,314)
\path(2767.000,434.000)(2797.000,314.000)(2827.000,434.000)
\put(900,936){\makebox(0,0)[lb]{\smash{{{\SetFigFont{10}{12.0}{\rmdefault}{\mddefault}{\itdefault}like}}}}}
\put(0,36){\makebox(0,0)[lb]{\smash{{{\SetFigFont{10}{12.0}{\rmdefault}{\mddefault}{\itdefault}john}}}}}
\put(1350,36){\makebox(0,0)[lb]{\smash{{{\SetFigFont{10}{12.0}{\rmdefault}{\mddefault}{\itdefault}woman}}}}}
\put(2700,36){\makebox(0,0)[lb]{\smash{{{\SetFigFont{10}{12.0}{\rmdefault}{\mddefault}{\itdefault}like}}}}}
\put(2700,936){\makebox(0,0)[lb]{\smash{{{\SetFigFont{10}{12.0}{\rmdefault}{\mddefault}{\itdefault}not}}}}}
\put(300,546){\makebox(0,0)[lb]{\smash{{{\SetFigFont{10}{12.0}{\rmdefault}{\mddefault}{\itdefault}1}}}}}
\put(1462,561){\makebox(0,0)[lb]{\smash{{{\SetFigFont{10}{12.0}{\rmdefault}{\mddefault}{\itdefault}2}}}}}
\put(2610,554){\makebox(0,0)[lb]{\smash{{{\SetFigFont{10}{12.0}{\rmdefault}{\mddefault}{\itdefault}1}}}}}
\end{picture}
}

%% file: D3.tex
\setlength{\unitlength}{0.00066667in}
\begingroup\makeatletter\ifx\SetFigFont\undefined%
\gdef\SetFigFont#1#2#3#4#5{%
  \reset@font\fontsize{#1}{#2pt}%
  \fontfamily{#3}\fontseries{#4}\fontshape{#5}%
  \selectfont}%
\fi\endgroup%
{\renewcommand{\dashlinestretch}{30}
\begin{picture}(3605,2709)(0,-10)
\path(975,2511)(375,1986)
\path(445.554,2087.598)(375.000,1986.000)(485.064,2042.443)
\path(1125,2511)(1725,1986)
\path(1614.936,2042.443)(1725.000,1986.000)(1654.446,2087.598)
\path(1800,1611)(1350,1086)
\path(1405.317,1196.635)(1350.000,1086.000)(1450.873,1157.587)
\path(1950,1611)(2700,1086)
\path(2584.488,1130.239)(2700.000,1086.000)(2618.896,1179.392)
\path(2700,786)(2475,261)
\path(2494.696,383.115)(2475.000,261.000)(2549.845,359.480)
\path(2850,786)(3375,261)
\path(3268.934,324.640)(3375.000,261.000)(3311.360,367.066)
\put(900,2586){\makebox(0,0)[lb]{\smash{{{\SetFigFont{10}{12.0}{\rmdefault}{\mddefault}{\itdefault}hate}}}}}
\put(0,1761){\makebox(0,0)[lb]{\smash{{{\SetFigFont{10}{12.0}{\rmdefault}{\mddefault}{\itdefault}peter}}}}}
\put(1575,1761){\makebox(0,0)[lb]{\smash{{{\SetFigFont{10}{12.0}{\rmdefault}{\mddefault}{\itdefault}woman}}}}}
\put(1125,861){\makebox(0,0)[lb]{\smash{{{\SetFigFont{10}{12.0}{\rmdefault}{\mddefault}{\itdefault}every}}}}}
\put(2625,861){\makebox(0,0)[lb]{\smash{{{\SetFigFont{10}{12.0}{\rmdefault}{\mddefault}{\itdefault}like}}}}}
\put(2325,36){\makebox(0,0)[lb]{\smash{{{\SetFigFont{10}{12.0}{\rmdefault}{\mddefault}{\itdefault}john}}}}}
\put(3375,36){\makebox(0,0)[lb]{\smash{{{\SetFigFont{10}{12.0}{\rmdefault}{\mddefault}{\itdefault}not}}}}}
\put(525,2286){\makebox(0,0)[lb]{\smash{{{\SetFigFont{10}{12.0}{\rmdefault}{\mddefault}{\itdefault}1}}}}}
\put(1575,2286){\makebox(0,0)[lb]{\smash{{{\SetFigFont{10}{12.0}{\rmdefault}{\mddefault}{\itdefault}2}}}}}
\put(1275,1311){\makebox(0,0)[lb]{\smash{{{\SetFigFont{10}{12.0}{\rmdefault}{\mddefault}{\itdefault}det}}}}}
\put(2400,1386){\makebox(0,0)[lb]{\smash{{{\SetFigFont{10}{12.0}{\rmdefault}{\mddefault}{\itdefault}-2}}}}}
\put(2400,486){\makebox(0,0)[lb]{\smash{{{\SetFigFont{10}{12.0}{\rmdefault}{\mddefault}{\itdefault}1}}}}}
\put(3225,486){\makebox(0,0)[lb]{\smash{{{\SetFigFont{10}{12.0}{\rmdefault}{\mddefault}{\itdefault}-1}}}}}
\end{picture}
}

%% file: AT2.tex
\setlength{\unitlength}{0.00066667in}
\begingroup\makeatletter\ifx\SetFigFont\undefined%
\gdef\SetFigFont#1#2#3#4#5{%
  \reset@font\fontsize{#1}{#2pt}%
  \fontfamily{#3}\fontseries{#4}\fontshape{#5}%
  \selectfont}%
\fi\endgroup%
{\renewcommand{\dashlinestretch}{30}
\begin{picture}(3617,3033)(0,-10)
\path(2850,1785)(2325,1110)
\path(2374.992,1223.140)(2325.000,1110.000)(2422.353,1186.304)
\path(2925,1785)(3375,1110)
\path(3283.474,1193.205)(3375.000,1110.000)(3333.397,1226.487)
\path(3375,810)(3375,210)
\path(3345.000,330.000)(3375.000,210.000)(3405.000,330.000)
\put(3225,30){\makebox(0,0)[lb]{\smash{{{\SetFigFont{10}{12.0}{\rmdefault}{\mddefault}{\itdefault}peter}}}}}
\put(2325,1485){\makebox(0,0)[lb]{\smash{{{\SetFigFont{10}{12.0}{\rmdefault}{\mddefault}{\itdefault}det}}}}}
\put(3300,1410){\makebox(0,0)[lb]{\smash{{{\SetFigFont{10}{12.0}{\rmdefault}{\mddefault}{\itdefault}-2}}}}}
\put(3525,510){\makebox(0,0)[lb]{\smash{{{\SetFigFont{10}{12.0}{\rmdefault}{\mddefault}{\itdefault}1}}}}}
\put(2625,1815){\makebox(0,0)[lb]{\smash{{{\SetFigFont{10}{12.0}{\rmdefault}{\mddefault}{\itdefault}woman}}}}}
\put(2025,938){\makebox(0,0)[lb]{\smash{{{\SetFigFont{10}{12.0}{\rmdefault}{\mddefault}{\itdefault}every}}}}}
\put(3225,893){\makebox(0,0)[lb]{\smash{{{\SetFigFont{10}{12.0}{\rmdefault}{\mddefault}{\itdefault}hate}}}}}
\path(1275,2835)(300,2160)
\path(381.587,2252.971)(300.000,2160.000)(415.739,2203.639)
\path(1350,2835)(1500,2160)
\path(1444.683,2270.635)(1500.000,2160.000)(1503.254,2283.650)
\path(1500,2835)(2775,2160)
\path(2654.909,2189.633)(2775.000,2160.000)(2682.982,2242.660)
\put(1200,2910){\makebox(0,0)[lb]{\smash{{{\SetFigFont{10}{12.0}{\rmdefault}{\mddefault}{\itdefault}like}}}}}
\put(600,2535){\makebox(0,0)[lb]{\smash{{{\SetFigFont{10}{12.0}{\rmdefault}{\mddefault}{\itdefault}1}}}}}
\put(0,1965){\makebox(0,0)[lb]{\smash{{{\SetFigFont{10}{12.0}{\rmdefault}{\mddefault}{\itdefault}john}}}}}
\put(1350,1883){\makebox(0,0)[lb]{\smash{{{\SetFigFont{10}{12.0}{\rmdefault}{\mddefault}{\itdefault}not}}}}}
\put(1125,2385){\makebox(0,0)[lb]{\smash{{{\SetFigFont{10}{12.0}{\rmdefault}{\mddefault}{\itdefault}-1}}}}}
\put(2400,2535){\makebox(0,0)[lb]{\smash{{{\SetFigFont{10}{12.0}{\rmdefault}{\mddefault}{\itdefault}2}}}}}
\end{picture}
}

%% file: AT4.tex
\setlength{\unitlength}{0.00066667in}
\begingroup\makeatletter\ifx\SetFigFont\undefined%
\gdef\SetFigFont#1#2#3#4#5{%
  \reset@font\fontsize{#1}{#2pt}%
  \fontfamily{#3}\fontseries{#4}\fontshape{#5}%
  \selectfont}%
\fi\endgroup%
{\renewcommand{\dashlinestretch}{30}
\begin{picture}(4071,3069)(0,-10)
\path(1275,2835)(300,2160)
\path(381.587,2252.971)(300.000,2160.000)(415.739,2203.639)
\path(1350,2835)(1500,2160)
\path(1444.683,2270.635)(1500.000,2160.000)(1503.254,2283.650)
\path(1500,2835)(2775,2160)
\path(2654.909,2189.633)(2775.000,2160.000)(2682.982,2242.660)
\path(2850,1785)(2325,1110)
\path(2374.992,1223.140)(2325.000,1110.000)(2422.353,1186.304)
\path(2925,1785)(3375,1110)
\path(3283.474,1193.205)(3375.000,1110.000)(3333.397,1226.487)
\path(3375,810)(3375,210)
\path(3345.000,330.000)(3375.000,210.000)(3405.000,330.000)
\put(1200,2910){\makebox(0,0)[lb]{\smash{{{\SetFigFont{10}{12.0}{\rmdefault}{\mddefault}{\itdefault}like(l1,l2)}}}}}
\put(0,1965){\makebox(0,0)[lb]{\smash{{{\SetFigFont{10}{12.0}{\rmdefault}{\mddefault}{\itdefault}john}}}}}
\put(1350,1883){\makebox(0,0)[lb]{\smash{{{\SetFigFont{10}{12.0}{\rmdefault}{\mddefault}{\itdefault}not(n1)}}}}}
\put(3225,30){\makebox(0,0)[lb]{\smash{{{\SetFigFont{10}{12.0}{\rmdefault}{\mddefault}{\itdefault}peter}}}}}
\put(2325,1485){\makebox(0,0)[lb]{\smash{{{\SetFigFont{10}{12.0}{\rmdefault}{\mddefault}{\itdefault}det}}}}}
\put(3300,1410){\makebox(0,0)[lb]{\smash{{{\SetFigFont{10}{12.0}{\rmdefault}{\mddefault}{\itdefault}-h2}}}}}
\put(3525,510){\makebox(0,0)[lb]{\smash{{{\SetFigFont{10}{12.0}{\rmdefault}{\mddefault}{\itdefault}+h1}}}}}
\put(2025,938){\makebox(0,0)[lb]{\smash{{{\SetFigFont{10}{12.0}{\rmdefault}{\mddefault}{\itdefault}every}}}}}
\put(3225,893){\makebox(0,0)[lb]{\smash{{{\SetFigFont{10}{12.0}{\rmdefault}{\mddefault}{\itdefault}hate(h1,h2)}}}}}
\put(2625,1890){\makebox(0,0)[lb]{\smash{{{\SetFigFont{10}{12.0}{\rmdefault}{\mddefault}{\itdefault}woman}}}}}
\put(1050,2385){\makebox(0,0)[lb]{\smash{{{\SetFigFont{10}{12.0}{\rmdefault}{\mddefault}{\itdefault}-n1}}}}}
\put(2325,2535){\makebox(0,0)[lb]{\smash{{{\SetFigFont{10}{12.0}{\rmdefault}{\mddefault}{\itdefault}+l2}}}}}
\put(450,2535){\makebox(0,0)[lb]{\smash{{{\SetFigFont{10}{12.0}{\rmdefault}{\mddefault}{\itdefault}+l1}}}}}
\end{picture}
}

%% file: D5.tex
\setlength{\unitlength}{0.00061242in}
\begingroup\makeatletter\ifx\SetFigFont\undefined%
\gdef\SetFigFont#1#2#3#4#5{%
  \reset@font\fontsize{#1}{#2pt}%
  \fontfamily{#3}\fontseries{#4}\fontshape{#5}%
  \selectfont}%
\fi\endgroup%
{\renewcommand{\dashlinestretch}{30}
\begin{picture}(7608,2551)(0,-10)
\path(5040,2524)(7335,229)
\path(5040,2524)(4140,1624)
\path(5940,1624)(5040,724)
\path(6390,1174)(5490,274)
\put(3780,1377){\makebox(0,0)[lb]{\smash{{{\SetFigFont{9}{10.8}{\familydefault}{\mddefault}{\updefault}Dn}}}}}
\put(4725,522){\makebox(0,0)[lb]{\smash{{{\SetFigFont{9}{10.8}{\familydefault}{\mddefault}{\updefault}D2}}}}}
\put(5130,27){\makebox(0,0)[lb]{\smash{{{\SetFigFont{9}{10.8}{\familydefault}{\mddefault}{\updefault}D1}}}}}
\put(7470,49){\makebox(0,0)[lb]{\smash{{{\SetFigFont{9}{10.8}{\familydefault}{\mddefault}{\updefault}H}}}}}
\put(4320,949){\makebox(0,0)[lb]{\smash{{{\SetFigFont{9}{10.8}{\familydefault}{\mddefault}{\updefault}...}}}}}
\put(5625,724){\makebox(0,0)[lb]{\smash{{{\SetFigFont{9}{10.8}{\familydefault}{\mddefault}{\updefault}L1}}}}}
\put(5085,1054){\makebox(0,0)[lb]{\smash{{{\SetFigFont{9}{10.8}{\familydefault}{\mddefault}{\updefault}L2}}}}}
\put(4140,1969){\makebox(0,0)[lb]{\smash{{{\SetFigFont{9}{10.8}{\familydefault}{\mddefault}{\updefault}Ln}}}}}
\path(1035,1714)(225,1084)
\path(1170,1669)(1350,1084)
\path(1260,1714)(2205,1084)
\path(2565,1759)(3375,1759)
\path(2805.000,1819.000)(2565.000,1759.000)(2805.000,1699.000)
\path(3135.000,1699.000)(3375.000,1759.000)(3135.000,1819.000)
\put(585,897){\makebox(0,0)[lb]{\smash{{{\SetFigFont{9}{10.8}{\familydefault}{\mddefault}{\updefault}...}}}}}
\put(1215,837){\makebox(0,0)[lb]{\smash{{{\SetFigFont{9}{10.8}{\familydefault}{\mddefault}{\updefault}D2}}}}}
\put(2115,837){\makebox(0,0)[lb]{\smash{{{\SetFigFont{9}{10.8}{\familydefault}{\mddefault}{\updefault}D1}}}}}
\put(0,837){\makebox(0,0)[lb]{\smash{{{\SetFigFont{9}{10.8}{\familydefault}{\mddefault}{\updefault}Dn}}}}}
\put(1890,1399){\makebox(0,0)[lb]{\smash{{{\SetFigFont{9}{10.8}{\familydefault}{\mddefault}{\updefault}L1}}}}}
\put(990,1279){\makebox(0,0)[lb]{\smash{{{\SetFigFont{9}{10.8}{\familydefault}{\mddefault}{\updefault}L2}}}}}
\put(270,1384){\makebox(0,0)[lb]{\smash{{{\SetFigFont{9}{10.8}{\familydefault}{\mddefault}{\updefault}Ln}}}}}
\put(1035,1894){\makebox(0,0)[lb]{\smash{{{\SetFigFont{9}{10.8}{\familydefault}{\mddefault}{\updefault}H}}}}}
\end{picture}
}

%% file: AT6.tex
\setlength{\unitlength}{0.00061242in}
\begingroup\makeatletter\ifx\SetFigFont\undefined%
\gdef\SetFigFont#1#2#3#4#5{%
  \reset@font\fontsize{#1}{#2pt}%
  \fontfamily{#3}\fontseries{#4}\fontshape{#5}%
  \selectfont}%
\fi\endgroup%
{\renewcommand{\dashlinestretch}{30}
\begin{picture}(7292,7203)(0,-10)
\path(2895,7176)(6915,3156)
\path(4560,5511)(240,2271)
\path(3696,6389)(2068,5168)
\path(2925,7172)(1297,5951)
\path(2294,3802)(5227,869)
\path(3351,2729)(303,443)
\path(1462,1274)(2400,336)
\put(900,5691){\makebox(0,0)[lb]{\smash{{{\SetFigFont{9}{10.8}{\rmdefault}{\mddefault}{\itdefault}john}}}}}
\put(1725,4941){\makebox(0,0)[lb]{\smash{{{\SetFigFont{9}{10.8}{\rmdefault}{\mddefault}{\itdefault}not(n1)}}}}}
\put(6600,2901){\makebox(0,0)[lb]{\smash{{{\SetFigFont{9}{10.8}{\rmdefault}{\mddefault}{\itdefault}like(l1,l2)}}}}}
\put(4950,666){\makebox(0,0)[lb]{\smash{{{\SetFigFont{9}{10.8}{\rmdefault}{\mddefault}{\itdefault}woman}}}}}
\put(0,2076){\makebox(0,0)[lb]{\smash{{{\SetFigFont{9}{10.8}{\rmdefault}{\mddefault}{\itdefault}every}}}}}
\put(1650,6591){\makebox(0,0)[lb]{\smash{{{\SetFigFont{9}{10.8}{\rmdefault}{\mddefault}{\itdefault}+l1}}}}}
\put(2475,5781){\makebox(0,0)[lb]{\smash{{{\SetFigFont{9}{10.8}{\rmdefault}{\mddefault}{\itdefault}-n1}}}}}
\put(3150,4716){\makebox(0,0)[lb]{\smash{{{\SetFigFont{9}{10.8}{\rmdefault}{\mddefault}{\itdefault}+l2}}}}}
\put(900,3096){\makebox(0,0)[lb]{\smash{{{\SetFigFont{9}{10.8}{\rmdefault}{\mddefault}{\itdefault}det}}}}}
\put(1905,2016){\makebox(0,0)[lb]{\smash{{{\SetFigFont{9}{10.8}{\rmdefault}{\mddefault}{\itdefault}-h2}}}}}
\put(570,966){\makebox(0,0)[lb]{\smash{{{\SetFigFont{9}{10.8}{\rmdefault}{\mddefault}{\itdefault}+h1}}}}}
\put(15,81){\makebox(0,0)[lb]{\smash{{{\SetFigFont{9}{10.8}{\rmdefault}{\mddefault}{\itdefault}peter}}}}}
\put(2085,36){\makebox(0,0)[lb]{\smash{{{\SetFigFont{9}{10.8}{\rmdefault}{\mddefault}{\itdefault}hate(h1,h2)}}}}}
\end{picture}
}

%% file: AT7.tex
\setlength{\unitlength}{0.00052493in}
\begingroup\makeatletter\ifx\SetFigFont\undefined%
\gdef\SetFigFont#1#2#3#4#5{%
  \reset@font\fontsize{#1}{#2pt}%
  \fontfamily{#3}\fontseries{#4}\fontshape{#5}%
  \selectfont}%
\fi\endgroup%
{\renewcommand{\dashlinestretch}{30}
\begin{picture}(8725,7203)(0,-10)
\path(3525,7176)(7545,3156)
\path(5190,5511)(870,2271)
\path(4326,6389)(2698,5168)
\path(3555,7172)(1927,5951)
\path(2924,3802)(5857,869)
\path(3981,2729)(933,443)
\path(2092,1274)(3030,336)
\put(1530,5691){\makebox(0,0)[lb]{\smash{{{\SetFigFont{8}{9.6}{\rmdefault}{\mddefault}{\itdefault}john: e}}}}}
\put(2355,4941){\makebox(0,0)[lb]{\smash{{{\SetFigFont{8}{9.6}{\rmdefault}{\mddefault}{\itdefault}not(n1): t}}}}}
\put(7230,2901){\makebox(0,0)[lb]{\smash{{{\SetFigFont{8}{9.6}{\rmdefault}{\mddefault}{\itdefault}like(l1,l2): t}}}}}
\put(5580,666){\makebox(0,0)[lb]{\smash{{{\SetFigFont{8}{9.6}{\rmdefault}{\mddefault}{\itdefault}woman: e\ra t}}}}}
\put(2280,6591){\makebox(0,0)[lb]{\smash{{{\SetFigFont{8}{9.6}{\rmdefault}{\mddefault}{\itdefault}+l1}}}}}
\put(3105,5781){\makebox(0,0)[lb]{\smash{{{\SetFigFont{8}{9.6}{\rmdefault}{\mddefault}{\itdefault}-n1}}}}}
\put(3780,4716){\makebox(0,0)[lb]{\smash{{{\SetFigFont{8}{9.6}{\rmdefault}{\mddefault}{\itdefault}+l2}}}}}
\put(1530,3096){\makebox(0,0)[lb]{\smash{{{\SetFigFont{8}{9.6}{\rmdefault}{\mddefault}{\itdefault}det}}}}}
\put(1200,966){\makebox(0,0)[lb]{\smash{{{\SetFigFont{8}{9.6}{\rmdefault}{\mddefault}{\itdefault}+h1}}}}}
\put(645,81){\makebox(0,0)[lb]{\smash{{{\SetFigFont{8}{9.6}{\rmdefault}{\mddefault}{\itdefault}peter: e}}}}}
\put(2715,36){\makebox(0,0)[lb]{\smash{{{\SetFigFont{8}{9.6}{\rmdefault}{\mddefault}{\itdefault}hate(h1,h2): t}}}}}
\put(2895,2286){\makebox(0,0)[lb]{\smash{{{\SetFigFont{8}{9.6}{\rmdefault}{\mddefault}{\itdefault}-h2}}}}}
\put(3210,426){\makebox(0,0)[lb]{\smash{{{\SetFigFont{8}{9.6}{\familydefault}{\mddefault}{\updefault}\so{h1:e, h2:e}}}}}}
\put(7710,3261){\makebox(0,0)[lb]{\smash{{{\SetFigFont{8}{9.6}{\familydefault}{\mddefault}{\updefault}\so{l1:e, l2:e}}}}}}
\put(3075,5196){\makebox(0,0)[lb]{\smash{{{\SetFigFont{8}{9.6}{\familydefault}{\mddefault}{\updefault}\so{n1:t}}}}}}
\put(0,2076){\makebox(0,0)[lb]{\smash{{{\SetFigFont{8}{9.6}{\rmdefault}{\mddefault}{\itdefault}every:  (e\ra t)\ra(e\ra t)\ra t}}}}}
\end{picture}
}

%% file: AT8.tex
\setlength{\unitlength}{0.00052493in}
\begingroup\makeatletter\ifx\SetFigFont\undefined%
\gdef\SetFigFont#1#2#3#4#5{%
  \reset@font\fontsize{#1}{#2pt}%
  \fontfamily{#3}\fontseries{#4}\fontshape{#5}%
  \selectfont}%
\fi\endgroup%
{\renewcommand{\dashlinestretch}{30}
\begin{picture}(6912,4132)(0,-10)
\path(1004,3253)(284,2353)
\path(1004,3253)(1724,2353)
\path(3434,3253)(2714,2353)
\path(3434,3253)(4154,2353)
\path(5909,3253)(5189,2353)
\path(5909,3253)(6629,2353)
\path(1004,1228)(284,328)
\path(1004,1228)(1724,328)
\path(3434,1228)(2714,328)
\path(3434,1228)(4154,328)
\path(5909,1228)(5189,328)
\path(5909,1228)(6629,328)
\put(1049,3973){\makebox(0,0)[b]{\smash{{{\SetFigFont{8}{9.6}{\familydefault}{\mddefault}{\updefault}complement}}}}}
\put(1049,3733){\makebox(0,0)[b]{\smash{{{\SetFigFont{8}{9.6}{\familydefault}{\mddefault}{\updefault}incorporation}}}}}
\put(3389,3973){\makebox(0,0)[b]{\smash{{{\SetFigFont{8}{9.6}{\familydefault}{\mddefault}{\updefault}modifier}}}}}
\put(3389,3733){\makebox(0,0)[b]{\smash{{{\SetFigFont{8}{9.6}{\familydefault}{\mddefault}{\updefault}incorporation}}}}}
\put(5909,3928){\makebox(0,0)[b]{\smash{{{\SetFigFont{8}{9.6}{\familydefault}{\mddefault}{\updefault}determiner}}}}}
\put(5909,3688){\makebox(0,0)[b]{\smash{{{\SetFigFont{8}{9.6}{\familydefault}{\mddefault}{\updefault}incorporation}}}}}
\path(959,1993)(959,1453)
\path(899.000,1693.000)(959.000,1453.000)(1019.000,1693.000)
\path(3434,1993)(3434,1453)
\path(3374.000,1693.000)(3434.000,1453.000)(3494.000,1693.000)
\path(5909,1993)(5909,1453)
\path(5849.000,1693.000)(5909.000,1453.000)(5969.000,1693.000)
\put(239,2083){\makebox(0,0)[b]{\smash{{{\SetFigFont{8}{9.6}{\familydefault}{\mddefault}{\updefault}\dprim}}}}}
\put(1589,2083){\makebox(0,0)[b]{\smash{{{\SetFigFont{8}{9.6}{\familydefault}{\mddefault}{\updefault}\hprimx}}}}}
\put(3974,2128){\makebox(0,0)[b]{\smash{{{\SetFigFont{8}{9.6}{\familydefault}{\mddefault}{\updefault}\hprim}}}}}
\put(329,36){\makebox(0,0)[b]{\smash{{{\SetFigFont{8}{9.6}{\familydefault}{\mddefault}{\updefault}\dprim}}}}}
\put(1589,36){\makebox(0,0)[b]{\smash{{{\SetFigFont{8}{9.6}{\familydefault}{\mddefault}{\updefault}\lhprim}}}}}
\put(2714,36){\makebox(0,0)[b]{\smash{{{\SetFigFont{8}{9.6}{\familydefault}{\mddefault}{\updefault}\ldprim}}}}}
\put(4109,36){\makebox(0,0)[b]{\smash{{{\SetFigFont{8}{9.6}{\familydefault}{\mddefault}{\updefault}\hprim}}}}}
\put(5144,36){\makebox(0,0)[b]{\smash{{{\SetFigFont{8}{9.6}{\familydefault}{\mddefault}{\updefault}\dprim}}}}}
\put(6674,36){\makebox(0,0)[b]{\smash{{{\SetFigFont{8}{9.6}{\familydefault}{\mddefault}{\updefault}\hprim}}}}}
\put(5144,2128){\makebox(0,0)[b]{\smash{{{\SetFigFont{8}{9.6}{\familydefault}{\mddefault}{\updefault}\dprim}}}}}
\put(6629,2128){\makebox(0,0)[b]{\smash{{{\SetFigFont{8}{9.6}{\familydefault}{\mddefault}{\updefault}\hprim}}}}}
\put(2669,2083){\makebox(0,0)[b]{\smash{{{\SetFigFont{8}{9.6}{\familydefault}{\mddefault}{\updefault}\dprimx}}}}}
\put(5324,2848){\makebox(0,0)[lb]{\smash{{{\SetFigFont{8}{9.6}{\familydefault}{\mddefault}{\updefault}det}}}}}
\put(419,2803){\makebox(0,0)[b]{\smash{{{\SetFigFont{8}{9.6}{\familydefault}{\mddefault}{\updefault}\px}}}}}
\put(2759,2758){\makebox(0,0)[lb]{\smash{{{\SetFigFont{8}{9.6}{\familydefault}{\mddefault}{\updefault}\mx}}}}}
\put(374,868){\makebox(0,0)[lb]{\smash{{{\SetFigFont{8}{9.6}{\familydefault}{\mddefault}{\updefault}\pp}}}}}
\put(2759,868){\makebox(0,0)[lb]{\smash{{{\SetFigFont{8}{9.6}{\familydefault}{\mddefault}{\updefault}\mm}}}}}
\put(5279,868){\makebox(0,0)[lb]{\smash{{{\SetFigFont{8}{9.6}{\familydefault}{\mddefault}{\updefault}det}}}}}
\end{picture}
}

%% file: AT9.tex
\setlength{\unitlength}{0.00048119in}
\begingroup\makeatletter\ifx\SetFigFont\undefined%
\gdef\SetFigFont#1#2#3#4#5{%
  \reset@font\fontsize{#1}{#2pt}%
  \fontfamily{#3}\fontseries{#4}\fontshape{#5}%
  \selectfont}%
\fi\endgroup%
{\renewcommand{\dashlinestretch}{30}
\begin{picture}(8076,7479)(0,-10)
\path(3525,7176)(6720,3936)
\path(5190,5511)(870,2271)
\path(4326,6389)(2698,5168)
\path(3555,7172)(1927,5951)
\path(2924,3802)(5857,869)
\path(3981,2729)(933,443)
\path(2092,1274)(3030,336)
\path(6990,3666)(7505,3156)
\put(1530,5691){\makebox(0,0)[lb]{\smash{{{\SetFigFont{7}{8.4}{\rmdefault}{\mddefault}{\itdefault}john: e}}}}}
\put(2355,4941){\makebox(0,0)[lb]{\smash{{{\SetFigFont{7}{8.4}{\rmdefault}{\mddefault}{\itdefault}not(n1): t}}}}}
\put(7230,2901){\makebox(0,0)[lb]{\smash{{{\SetFigFont{7}{8.4}{\rmdefault}{\mddefault}{\itdefault}like(l1,l2): t}}}}}
\put(5580,666){\makebox(0,0)[lb]{\smash{{{\SetFigFont{7}{8.4}{\rmdefault}{\mddefault}{\itdefault}woman: e\ra t}}}}}
\put(2280,6591){\makebox(0,0)[lb]{\smash{{{\SetFigFont{7}{8.4}{\rmdefault}{\mddefault}{\itdefault}+l1}}}}}
\put(3105,5781){\makebox(0,0)[lb]{\smash{{{\SetFigFont{7}{8.4}{\rmdefault}{\mddefault}{\itdefault}-n1}}}}}
\put(3780,4716){\makebox(0,0)[lb]{\smash{{{\SetFigFont{7}{8.4}{\rmdefault}{\mddefault}{\itdefault}+l2}}}}}
\put(1530,3096){\makebox(0,0)[lb]{\smash{{{\SetFigFont{7}{8.4}{\rmdefault}{\mddefault}{\itdefault}det}}}}}
\put(1200,966){\makebox(0,0)[lb]{\smash{{{\SetFigFont{7}{8.4}{\rmdefault}{\mddefault}{\itdefault}+h1}}}}}
\put(645,81){\makebox(0,0)[lb]{\smash{{{\SetFigFont{7}{8.4}{\rmdefault}{\mddefault}{\itdefault}peter: e}}}}}
\put(2715,36){\makebox(0,0)[lb]{\smash{{{\SetFigFont{7}{8.4}{\rmdefault}{\mddefault}{\itdefault}hate(h1,h2): t}}}}}
\put(2895,2286){\makebox(0,0)[lb]{\smash{{{\SetFigFont{7}{8.4}{\rmdefault}{\mddefault}{\itdefault}-h2}}}}}
\put(0,2076){\makebox(0,0)[lb]{\smash{{{\SetFigFont{7}{8.4}{\rmdefault}{\mddefault}{\itdefault}every:  (e\ra t)\ra(e\ra t)\ra t}}}}}
\put(3390,6681){\makebox(0,0)[lb]{\smash{{{\SetFigFont{7}{8.4}{\familydefault}{\mddefault}{\updefault}C2}}}}}
\put(4200,5961){\makebox(0,0)[lb]{\smash{{{\SetFigFont{7}{8.4}{\familydefault}{\mddefault}{\updefault}C4}}}}}
\put(5055,5061){\makebox(0,0)[lb]{\smash{{{\SetFigFont{7}{8.4}{\familydefault}{\mddefault}{\updefault}C1}}}}}
\put(2805,3441){\makebox(0,0)[lb]{\smash{{{\SetFigFont{7}{8.4}{\familydefault}{\mddefault}{\updefault}C3}}}}}
\put(3840,2361){\makebox(0,0)[lb]{\smash{{{\SetFigFont{7}{8.4}{\familydefault}{\mddefault}{\updefault}C5}}}}}
\put(1950,921){\makebox(0,0)[lb]{\smash{{{\SetFigFont{7}{8.4}{\familydefault}{\mddefault}{\updefault}C2}}}}}
\put(2535,1191){\makebox(0,0)[lb]{\smash{{{\SetFigFont{7}{8.4}{\familydefault}{\mddefault}{\updefault}\hateone}}}}}
\put(5055,5691){\makebox(0,0)[lb]{\smash{{{\SetFigFont{7}{8.4}{\familydefault}{\mddefault}{\updefault}\likeone}}}}}
\put(4155,6636){\makebox(0,0)[lb]{\smash{{{\SetFigFont{7}{8.4}{\familydefault}{\mddefault}{\updefault}\liketwo}}}}}
\put(3120,7356){\makebox(0,0)[lb]{\smash{{{\SetFigFont{7}{8.4}{\familydefault}{\mddefault}{\updefault}\likethree}}}}}
\put(3120,3756){\makebox(0,0)[lb]{\smash{{{\SetFigFont{7}{8.4}{\familydefault}{\mddefault}{\updefault}\womantwo}}}}}
\put(4200,2586){\makebox(0,0)[lb]{\smash{{{\SetFigFont{7}{8.4}{\familydefault}{\mddefault}{\updefault}\womanone}}}}}
\end{picture}
}